\documentclass{article}

\newcommand{\paperTitle}{MiDAS: Multi-integrated Domain Adaptive Supervision for Fake News Detection}

\PassOptionsToPackage{round, sort}{natbib}
\usepackage[preprint]{neurips_2022}

\usepackage[utf8]{inputenc} % allow utf-8 input
\usepackage[T1]{fontenc}    % use 8-bit T1 fonts
\usepackage{hyperref}       % hyperlinks
\usepackage{url}            % simple URL typesetting
\usepackage{booktabs}       % professional-quality tables
\usepackage{amsfonts}       % blackboard math symbols
\usepackage{nicefrac}       % compact symbols for 1/2, etc.
\usepackage{microtype}      % microtypography
\usepackage[table,xcdraw]{xcolor}
\usepackage{xspace}
\usepackage{multirow}
\usepackage{amsmath,amsopn}
\usepackage{xstring}
\usepackage{breqn}
\usepackage[capitalize,noabbrev,nameinlink]{cleveref}
\usepackage{dsfont}
\usepackage{graphicx}
\usepackage{floatrow}
\usepackage{blindtext}
\usepackage[font=small,labelfont=bf,tableposition=top]{caption}
\usepackage{subcaption}
\usepackage{varwidth}
\usepackage{multirow}
\usepackage[normalem]{ulem}
\usepackage{booktabs}

\newcommand{\PP}[1]{
    \vspace{2px}
    \noindent{\bf \IfEndWith{#1}{.}{#1}{#1.}}
    }

\newcommand{\squishitemize}{
\begin{list}{$\bullet$}
	{ \setlength{\itemsep}{0pt}
		\setlength{\parsep}{3pt}
		\setlength{\topsep}{3pt}
		\setlength{\partopsep}{0pt}
		\setlength{\leftmargin}{1.95em}
		\setlength{\labelwidth}{1.5em}
		\setlength{\labelsep}{0.5em} } }

\newcounter{Lcount}
\newcommand{\squishlist}{
	\begin{list}{\arabic{Lcount}. }
		{ \usecounter{Lcount}
			\setlength{\itemsep}{0pt}
			\setlength{\parsep}{3pt}
			\setlength{\topsep}{3pt}
			\setlength{\partopsep}{0pt}
			\setlength{\leftmargin}{2em}
			\setlength{\labelwidth}{1.5em}
			\setlength{\labelsep}{0.5em} } }
	
\newcommand{\squishend}{\end{list}}

\newcommand{\ttt}{\texttt}

\newcommand{\sys}{\textsc{MiDAS}\xspace}

\newcommand{\ie}{\textit{i}.\textit{e}.\xspace}

\DeclareMathOperator*{\argmin}{arg\,min}

\newfloatcommand{capbtabbox}{table}[][\FBwidth]
\useunder{\uline}{\ul}{}

\title{\paperTitle}

\author{%
  Abhijit Suprem \\
  School of Computer Science\\
  Georgia Institute of Technology\\
  Atlanta, GA 30313 \\
  \texttt{asuprem@gatech.edu} \\
  \And
  Calton Pu \\
  School of Computer Science\\
  Georgia Institute of Technology\\
  Atlanta, GA 30313 \\
  \texttt{calton.pu@cc.gatech.edu} \\
}

\begin{document}

\maketitle

\begin{abstract}
Covid-19 related misinformation and fake news, coined an `infodemic',
has dramatically increased over the past few years.  
This misinformation exhibits concept drift, where the 
distribution of fake news changes over time, reducing
effectiveness of previously trained models for fake news detection.
Given a set of fake news models trained on multiple domains,
we propose an adaptive decision module to select the best-fit model 
for a new sample.
We propose \sys, a multi-domain adaptative approach for  
fake news detection that ranks relevancy of existing
models to new samples. 
\sys contains 2 components: a doman-invariant encoder, 
and an adaptive model selector.
\sys integrates multiple pre-trained and fine-tuned models with 
their training data to create a domain-invariant representation. 
Then, \sys uses local Lipschitz smoothness of the invariant embedding 
space to estimate each model's relevance to a new sample.
Higher ranked models provide predictions, and lower ranked models
abstain.
We evaluate \sys on generalization to drifted data with 9 fake news datasets, 
each obtained from different domains and modalities.
\sys achieves new state-of-the-art performance on multi-domain adaptation 
for out-of-distribution fake news classification.  
\end{abstract}
  
\section{Introduction}
\label{sec:intro}
%Misinformation is a staple of open communication such as social media [ref]. 
%
The misinformation and fake news associated with the COVID-19 pandemic, 
called an `infodemic' by WHO \citep{infodemic}, have grown dramatically, 
and evolved with the pandemic. 
%
% Instead of harmless curiosity, fake news are believed \citep{misinfo, infodemic} 
% to have significant negative impact on the many efforts to manage and 
% reduce the pandemic, e.g., the anti-vaccination campaigns. 
%
Fake news has eroded institutional trust \citep{infotrust}
and have increasingly negative impacts outside social communities \citep{infolife}. 
%
% For the pandemic and decisive events such as elections, 
% the importance of detecting and filtering active fake news campaigns 
% cannot be overstated. 
%
The challenge is to filter active fake news campaigns 
while they are raging, just like today's online email spam filters, 
instead of offline, retrospective detection long after the 
campaigns have ended. 
We divide this challenge to detect fake news online into two 
parts: (1) the variety of data (both real and fake), and 
(2) the timeliness of data collection and processing (both real and fake). 
In this paper, we focus on the first (variety) part of challenge, 
with the timeliness (which depends on solutions to handle variety) in future work \citep{ebka}.

The infodemic, and fake news more generally, evolves with 
a growing variety of ephemeral topics and content, 
a phenomenon called real concept drift \citep{gamadrift}. 
%
% Previous attempts to handle the variety include: fake news 
% from multiple sources \citep{mdaws}, such as social media, 
% news articles, health blogs, and posts. 
%
However, the excellent results on single-domain classification 
\citep{fnft}, have generalization difficulties when applied to 
cross-domain experiments \citep{generalization, kmp}. 
A benchmark study over 15 language models shows reduced 
cross-domain fake news detection accuracy \citep{generalization}. 
A generalization study in \citep{kmp} finds significant 
performance deterioration when models are used on unseen, 
non-overlapping datasets. 
Intuitively, it is entirely reasonable that state-of-the-art 
models trained on one dataset or time period will have reduced 
accuracy on future time periods. 
%
% As concrete examples, models trained on 2016 elections are not 
% expected to work well for 2022 elections, and models trained 
% with 2020 COVID infodemic data would be unaware of the omicron 
% variant, discovered in November 2021, or the ivermectin conspiracy
% theory from Summer of 2021. 
%
Real concept drift is introduced into fake news as content changes 
\citep{gamadrift}, camouflage \citep{camo}, linguistic drift \citep{lexdiff}, 
and adversarial adaptation by fake news producers when faced 
with debunking efforts such as CDC on the pandemic \citep{misinfo1}.

To catch up with concept drift, the classification models need to 
be expanded to cover a wide variety of data sets \citep{mdaws, kmp, mcnnet}, or 
augmented with new knowledge on true novelty such as the appearance of 
the Omicron variant \citep{ebka}. 
%
% The expansion and augmentation are typically managed as an 
% ensemble or team of expert models that cover different time periods \citep{ebka}. 
% %
% Unlike typical ML models trained from data sets labeled by a 
% variety of methods such as crowdsourcing, fake news classifiers 
% require high level of trust, e.g.,  CDC situation reports \citep{cdcreport} 
% on the pandemic, and PolitiFact \citep{politifact}. 
%
In this paper, we assume the availability of domain-specific 
authorative sources such as CDC and WHO that provide 
trusted up-to-date information on the pandemic.
% , enabling the 
% augmentation of a teamed classifier with new models capable of 
% catching up with real concept drift. 

A key challenge of such multi-domain classifiers is a decision module 
to select the best-fit model amongst a set of existing models to 
classify new samples.
%  that may contain previously unseen text/words, 
% and make decisions on whether they are true novelty or fake news.  
%
% In some specific domains, steady and unidirectional drift may allow 
% simple heuristics to work, e.g., newer models perform better than 
% older models that have drifted longer and further. 
%
% In general, drift may take twist and turns, and we need to find the 
% model that `knows best' the new samples. 
%
This degree of knowledge is defined by the overlap between an 
unlabeled sample and existing models' training datasets \citep{kmp}. 
%
% The decision module gives higher weight to models that 
% show higher relevance, i.e., the amount of overlap between the 
% new samples and model training data, utilizing a variety of methods 
% to combine votes, such as team-of-experts [ref], weak supervision [ref], and ensembling [ref].
% A key challenge, then, is to select the best-fit model 
% amongst a set of older models to classify an unlabeled,
% potentially drifted sample. 
%
Intuitively, a best-fit model better captures a sample point's 
neighborhood in its own training data \cite{nice, liger}.
%
% So, we first calculate the overlap between an 
% unlabeled sample and existing training datasets, then
% rank models on relevance in training data to 
% the unlabeled sample.
% %
% In case of significant irrelevancy, a model should abstain 
% from predictions.
% %
% In case of multiple model relevancies, 
% any number of label integration methods can be used to combine votes, 
% such as team-of-experts \cite{ebka}, weak supervision \cite{snorkel}, or ensembling.

\PP{MiDAS.} 
We propose \sys, a multi-domain adaptative approach 
for early fake news detection, with potential for online filtering.
\sys integrates multiple pre-trained and fine-tuned models 
along with their training data to create a domain-invariant 
representation. 
On this representation, \sys uses a notion of local Lipschitz 
smoothness to estimate the overlap, and therefore relevancy, between 
a new sample and model training datasets.
This overlap estimate is used to rank models on 
relevancy to the new sample. 
Then, \sys selects the highest ranked model to perform classification. 
We evaluate \sys on 9 fake news datasets  obtained from different 
domains and modalities. 
We show new state-of-the-art performance on multi-domain 
adaptation for early fake news classification. 

\PP{Contributions.}
Our contributions are as follows:
\squishlist
\item \sys: a framework for adaptive model selection by using sample-to-data overlap to measure model relevancy
\item Experimental results of \sys on 9 fake news datasets with state-of-the-art results using unsupervised domain adaptation. 
\squishend
\section{Related Work}
\label{sec:related}

\subsection{Multi-Domain Adaptation}
Domain adaptation maps a target domain into a source domain. 
This allows a classifier learned from the source domain to 
predict the target domain samples \citep{domains}. 
Some approaches focus on a domain invariant representation 
between source and target \citep{coviddomains}. 
Then, a new classifier can be trained on this invariant 
representation for both source and target samples. 
Domain invariance is scalable to multiple source domains 
by fusing their latent representations with an 
adversarial encoder-discriminator framework \citep{mdaws}. 
For multi-source domain adaptation (MDA), classifiers for each 
source have different weights: static weights using 
distance \citep{mdaws} or per-sample weights on l2 norm \citep{odin}.

\subsection{Label Confidence}
Alongside domain adaptation, weak supervision (WS) is also common 
for propagating labels from source domains to a target domain \citep{snorkel}. 
Both approaches estimate labels closest to the true label 
of the target domain sample. 
This works with the assumption that the source domains 
or labeling functions, respectively, are correlated to 
the true labels due to expertise and domain knowledge. 
In each case, whether MDA or WS, domains or labeling 
functions need to be weighted to ensure reliance on 
the best-fit source. 
Snorkel, from \citep{snorkel}, uses expert labeling functions and weighs 
them on conditional independence. 
Similarly, approaches in \citep{epoxy, fast} use coverage of expert foundation models 
and weigh on distance to embedded sample. 
EEWS from \citep{eews} directly combines source data and labeling 
function in estimator parametrization to generate 
dynamic weights for each sample. 
MDA approaches weigh sources with weak supervision 
\citep{mdaws}, distance \citep{kmp}, or as team-of-experts \citep{ebka}.

% With MiDAS, we will compute label confidence by 
% directly using the embedding space of the source classifiers. 
% %
% Smoothness in the embedding space corresponds to 
% lift in labeling, per work on Lipschitzness in \citep{nice, liger}. 
% %
% With MiDAS, we use local smoothness of each source 
% classifier, computed from a domain-invariant representation, 
% to rank sources on embedding proximity to the unlabeled target sample.

\section{Problem Setup and Strategy}
\label{sec:problem}
Let there be $k$ source data domains, with
labels $\{X_i, Y_i\}_{i=1}^k \in \{D\}_(i=1)^k$.
Each of these source has an associated source model
SM, with a total of $k$ SMs: $\{f_i\}_(i=1)^k$, where
we have access to the training data $X_i$ and weights $w_i$.
Each SM yields hidden embeddings through a feature extractor
backbone, or foundation model \citep{foundation}.
Embeddings are projected to class probabilities with any type of 
classification layer/module.

\sys adaptively weights the $k$ SM predictions for some unlabeled,
potentially drifted target data domain $X'$.
We accomplish this by using local embedding smoothness of the SMs as a 
proxy for model relevance to a sample $x'\in X'$. 
SMs are typically smooth in the embedding space \citep{nice}; further, smoothness 
is correlated with local accuracy \citep{liger}.

With \sys, we rank each $f_k$ on a smoothness measure around the embedding 
for $x'$, \ie $f_k (x')$. 
Then, \sys can directly use the top-ranked $f_k$, or the smoothest $f_k$ under
a smoothness threshold, as the best-fit 
relevant models for $x'$, with the remaining models abstaining.
Because we are directly measuring smoothness on the embedding space, 
\sys can use already fine-tuned, state-of-the-art classifiers for each task, 
allowing off-the-shelf, plug-and-play usage. 
These classifiers are foundation models \citep{foundation} 
that have been fine-tuned with architectural changes, learned weights,
and hyperparameter tuning for their specific dataset.

There are two key challenges in \sys:
\squishlist
\item How do we compare smoothness of SMs that have been trained on different domains?
\item How can be measure the smoothness itself for unlabeled samples in the embedding space of SMs?
\squishend

We address (1) with an encoder $E$ that generates a domain invariant representation on
$\{X_i\}_{i=1}^k$. 
This unifies the data domains, allowing comparisons for different SMs to start 
from the same source domain.
For (2), we extend the idea of local Lipschitz smoothness from \citep{liger} and \citep{nice} 
to randomized Lipschitz smoothness.
In randomized Lipschitz smoothness, we randomly perturb $E(x')$,
the domain invariant representation of $x'$. 
Then, we compute the local
lipschitz constant $L$ on these perturbations $E(x')+\epsilon$, with respect to 
$E(x')$ to measure smoothness.
This allows us to calculate an $L_k$ for each $f_k$ and use the local Lipschitz constant,
a measure for the local smoothness, as a measure of relevancy. 
\section{MiDAS}
\label{sec:midas}

\begin{figure}
    \centering
    \includegraphics[width=\textwidth]{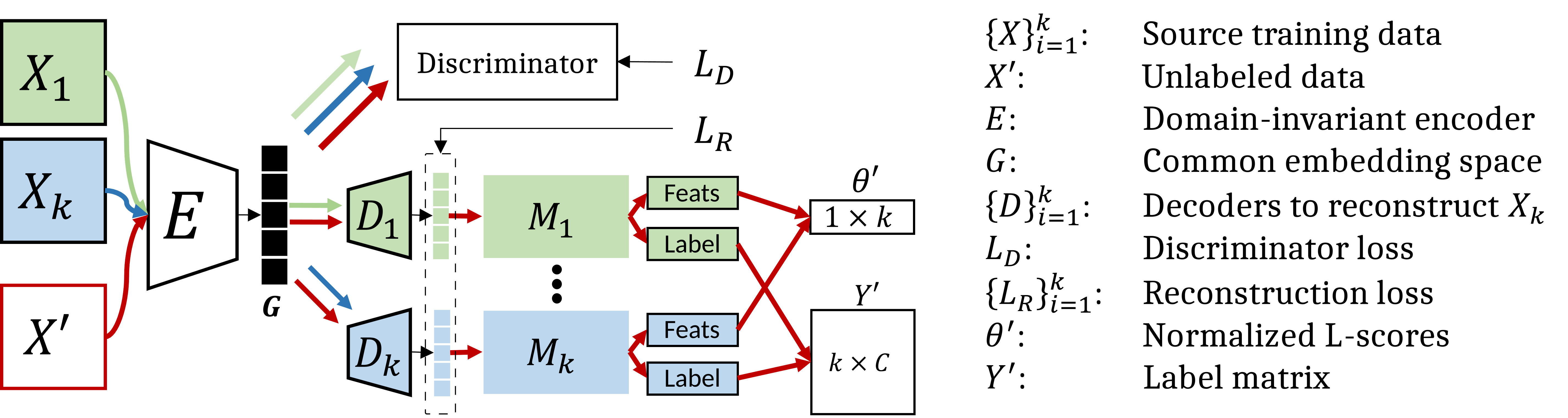}
    %\fbox{\rule[-.5cm]{0cm}{4cm} \rule[-.5cm]{4cm}{0cm}}
    \caption{\sys architecture: The encoder generates a domain invariant representation. We add perturbation to this representation. Then, fine-tuned models $\{M\}_(i=1)^k$ process the sample and perturbations. After computing the local Lipschitsz constant for each model, we can rank their L scores and select the best-fit model's label from the label matrix. }
    \label{fig:midas}
  \end{figure}

We now describe \sys components and implementation details. 
The \sys architecture is shown in \cref{fig:midas}. 
First, we cover the encoder-decoder framework to generate the 
domain invariant representations of source and target datasets. 
Then we present the randomized Lipschitz smoothness measure to generate 
SM relevancy rankings.

\subsection{Domain Invariant Encoder}
To compare different $f_k$ relevancy, we require a common source domain \citep{mdaws}.
We achieve this with a single-encoder multiple-decoders design,
where we have a single encoder to generate a domain invariant representation
from all source domains.
Then we use $k$ decoders to reconstruct the invariant representation for each 
$f_k$.
To train $E$, we use an adversarial discriminator $D'$ to enforce 
invariance with a min-max game, where the discriminator tries to
identify the source domain of the invariant representation,
and the encoder tries to fool it:

\begin{equation}
\label{eq:minmax}
\min_{E} \max_{D'} -\sum_{i=1}^k \mathds{E}_{(x,y)\sim X_k} [ l (D' (E (x)), k)  ] 
%- \mathds{E}_{x\sim X'} [ l (D' (E (x)), 0)  ]
\end{equation}

We use a gradient reversal layer $R$ to convert
the min-max to single-step minimization.
$R$ is the identity matrix during the forward pass,
and the discriminator gradient during the backward 
pass, scaled by $\lambda=-1$. 
Then, the adversarial optimization becomes:

\begin{equation}
    \label{eq:advloss}
    Loss = \min_{E, R(D')} -\sum_{i=1}^k \mathds{E}_{(x,y)\sim X_k} [ l (D' ( R(E (x))), k)  ] 
    %- \mathds{E}_{x\sim X'} [ l (D' (E (x)), 0)  ]
\end{equation}

To train $D_k$, we use the masked language modeling loss from BERT and AlBERT 
pretraining. 
Parameters are initialized with AlBERT's \ttt{albert-base-uncased} weights \citep{albert}.

In summary, we train a single encoder and $k$ decoders.
The encoder projects our training data in the form of 
SentencePiece tokens \citep{sentencepiece} into a domain-invariant representation,
trained with a domain discriminator.
Each decoder then reconstructs the original input training tokens from the
invariant representation.
We use decoders because \sys is designed to work with our existing BERT and AlBERT
classifiers, which expect SentencePiece token input.
Decoders are trained with masked language modeling, where we randomly mask
up to 15\% of words or tokens in the input.
Then, during prediction, an unlabeled, potentially drifted sample $x'$ from 
an unseen distribution $X'$ is converted to a domain invariant representation
$E(x')$.
Each decoder $D_k$ reconstructs from this invariant representation the input to 
its corresponding SM $f_k$.

\subsection{Randomized Lipschitz Smoothness}
\label{sec:rls}
With a common embedding space, we can now compare relevancy 
of each model to an unlabeled, potentially drift sample $x'$.
To present our approach, we need to introduce Lipschitz continuity.

\PP{Definition (Lipshitz Continuity)}
A function $f:\mathds{R}^n \rightarrow \mathds{R}^m$ is Lipschitz continuous if,
for some metric space $(X,\theta)$, there exists a constant $L$ such that

$$
\theta(f(x_1), f(x_2)) \leq L\cdot \theta(x_1, x_2)
$$

We can extend this to define Lipschitz-smooth with respect 
to SM predictions using Lipschitzness from \citep{liger}.

\PP{Definition (Lipshitzness)}
An SM is Lipschitz smooth if, for some class label $C$, 

\begin{equation}
\label{eq:lipschitzness}
|\Pr ( f_k (x_1) = C) - Pr ( f_k (x_2) = C)| \leq L_k \theta (x_1, x_2)
\end{equation}

That is, with $L_k$ smoothness, the diference in $f_k$'s predicted 
labels on $x_1$ and $x_2$ is bounded by $L_k$ for all $x\in X$.
However, the local value of $L_k$ can vary across the embedding
space.
Consequently, $f_k$ is smoother wherever $L_k$ is smaller 
\footnote{As $L_k\rightarrow 0$, the embedding function approaches mode
collapse, where every input point is projected to the same 
embedding point}.
As such, we want small $L_k$ for samples in the same class, and large $L_k$
for samples from different classes.% \citep{liger}.

With these definitions, we can present our approach for 
finding the best-fit relevant $f_k$ for $x'$, defined as the $f_k$ with
the smoothest embedding space around $x'$.

\PP{Theorem 1}
Let the best-fit $f_k$ for a sample $x'$ be the SM that is 
smoothest around $x'$.
We can find the best-fit $f_k$ for a particular sample $x'$,
given a distance threshold $\epsilon$, by solving:

\begin{equation}
\label{eq:theorem1}
\argmin_{k} \max_{\theta(x',x_r)\leq \epsilon|_{r=1}^N} \frac{\theta(\Pr (f_k(x') ), \Pr (f_k (x_r)))}{\theta(x', x_r)}
\end{equation}

The $\max$ term estimates the $L_k$ value for each $f_k$ by sampling $N$
random points in an $\epsilon$-Ball around $x'$.
Then, we find the $f_k$ that has the smallest
$L_k$.

A key insight is that adversarial attacks exploit non-smoothness
of a model's embedding space to fool classifiers,
by generating a noise $\epsilon$ such that $f_k(x+\epsilon)\neq f_k(x)$.
This non-smoothness occurs when $f_k$ does not capture enough training data 
in the region around $x$ properly;
in GANs, this causes `holes' in the latent space \citep{odin} during image
synthesis.
Conversely, adversarial defenses either enforce smoothness around embedding space 
or on potentially perturbed inputs themselves \citep{shield}. 
Similarly, GANs can enforce 1-Lipschitzness to improve coverage of 
sample generation \citep{ganlip} . 

So, given several $f_k$ with different local $L_k$ around the embedding
$f_k(x + \epsilon)$, a lower $L_k$ indicates smoother
embedding space \textit{because} that SM has captured more
training data in the region surrounding $x$ relative to other SMs,
similar to the overlap metric calculated in \citep{kmp}
However, even with a low $L_k$, the classification
labels $y_k = f'_k(f_k(x))$, obtained from the classification module 
$f'_k$ of the $k$th SM, can change on perturbations around $x$.
We can use probabilistic Lipschitzness to bound the probability 
of $y_k$ changing in the neighborhood $x+\epsilon$ as a function
of the perturbation $\epsilon$.

\PP{Definition (Probabilistic Lipschitzness)}
Let $\phi:\mathds{R}\rightarrow [0,1]$. Given $x',x_r\sim P_X$,
we say that $f_k$ is $\phi$-Lipschitz if, for all $\epsilon>0$,
there is an increasing function $\phi(\epsilon)$ such that:

\begin{equation}
\label{eq:pl}
\Pr_{x',x_r\sim P_X} [ \theta ( f_k (x'), f_k (x_r)) - \frac{1}{\epsilon} \theta ( x', x_r) > 0    ] \leq \phi ( \epsilon)
\end{equation}

That is, a function that is Lipschitz by Definition 1 (L-Lipschitz) satisfies 
Definition 3 ($\phi$-Lipschitz) with $\phi(\epsilon) = 1$ if $\epsilon \geq 1/L$
and $\phi (\epsilon)=0$ if $\epsilon < 1/L$.

From this, it follows that if $f_k$ satisfies the $\phi$-Lipschitz condition, then
the number of samples within an $\epsilon$-Ball of $x'$ that have a different
label from $x'$ is bounded by $\phi(\epsilon)$, per \citep{nice}.
As we move further from $x'$, the probability of a label change increases.
Let $a'$ be the accuracy for $f_k$ at $x'$. 
If we know the accuracy
drop $\alpha$ at the edge of the $\epsilon$-Ball where labels change values, 
we can bound the accuracy of predictions between $x'$ and some perturbed point 
$x_r$ in an $\epsilon$-Ball around $x'$ to:

$$
a_r\geq a' - \alpha\cdot\phi(\epsilon)
$$

If we calculate $\alpha$ for an SM using training examples of
different labels within the margins allowed by probabilistic
Lipschitzness, we find that accuracy bound depends only
on choice of $\epsilon$. 
Consequently we can approximate $f_k$'s accuracy on 
$x'$ if $x_r$ is within a small $\epsilon$-Ball around
$x'$ to be at least:

\begin{equation}
\label{eq:accbound}
\theta ( \Pr ( f_k ( x')), \Pr ( f_k (x_r))) \geq \Pr(f_k (x')=y_k) - \alpha\cdot\phi(\epsilon)
\end{equation}

\subsection{Putting it Together: MiDAS}
\label{sec:midasimpl}
Recall that the encoder projects all samples 
to a common invariant domain $E:\{X\}_{i=1}^k \rightarrow G$.
Each SM-specific decoder $D_k$ then converts from the invariant
domain to the $k$-th source domain $D_k:G\rightarrow X_k$.
With this encoder-decoders framework, we can use the invariant domain
$G$ as the common source domain for all SMs.

Per Definition 3, accuracy is best estimated in an $\epsilon<1\L$ ball.
Once $E$ is trained, we can use the class cluster centers 
(obtained from K-Means clustering on the embeddings) from
the training data to compute a local cluster-specific $L$ for
each cluster in each $f_k$.
This partitioning allows us to compute local smoothness
characteristics of the embedding space, simialr to the concurrent work
in \citep{liger}. 
We use the cluster centers because we want the strongest
measure of smoothness for each $f_k$, and this occurs near the cluster
center; for example \citep{kmp} and \citep{liger} both use the centroid 
to set up accuracy thresholds.
Then, we can estimate a local $L_k$ for each $f_k$ at the cluster centers
with the $m$ nearest point to the center using \cref{eq:lipschitzness}.
We take the maximum $L_k$ among all $f_k$ to obtain the upper bound on the 
local smoothness among the SMs.
On this maximum, we can calculate $\epsilon = 1 / \max({L_k}_(i=1)^k)$. 
So, the only hyperparameter here is $m$, the number of neighbors to compute
the local $L_k$ for each label cluster in each $f_k$.
We explore impact of $m$ in \cref{sec:meval}.

Now, for an unlabeled sample $x'$, we first generate the domain invariant
representation $x'_G = E(x')$.
Then we perturb $x'_G$ to generate $r$ points $\{x'_Gi \}_{i=1}^r$ in an $\epsilon$-Ball around $x'_G$.
Using $x'_G$ and $\{x'_Gi \}_{i=1}^r$, we can compute the local
Lipschitz constant $L'_k$ for each $f_k$ using \cref{eq:lipschitzness}.
We have 2 possibilities:
\squishlist
\item $L'_k \geq 1/\epsilon$: $f_k$ that satisfy this condition 
abstain from providing predictions, since the accuracy drop is unbounded.
\item $L'_k < 1/\epsilon$: $f_k$ that satisfy this condition
can provide labels, because their accuracy is bounded
per \cref{eq:accbound}.
\squishend
\section{Evaluation}
\label{sec:eval}

\subsection{Experimental Setup and Datasets}
We implemented and evaluated \sys on PyTorch 1.11 
on a server running NVIDIA T100 GPUs. We have released our implementation code.

% \usepackage{booktabs}
\begin{table}[]
    \caption{We use 9 COVID-19 fake news datasets to evaluate MiDAS. Here we also present a motivating experiment with respect to concept drift and generalizability: for each dataset, we train an `oracle' on the training data. The oracle's performance on the test data is compared to an ensemble of the other 8 datasets. The latter tests the concept drift case, where models need to generalize to data distributions they have not yet encountered. There is significant accuracy drop due to concept drift, domain shift, and label overlap}
    \label{tab:datasets}
    \small
    \begin{tabular}{@{}lrrrrr@{}}
    \toprule
    Dataset       & Training & Testing & \begin{tabular}[c]{@{}r@{}}Oracle Acc.\\ (Fine-Tuning)\end{tabular} & \begin{tabular}[c]{@{}r@{}}Generalization Acc.\\ (Held-Out)\end{tabular} & Decrease \\ \midrule
    kaggle\_short \citep{kagglefn} & 31k      & 9K      & 0.97                                                                & 0.57                                                                     & 42\%     \\
    kaggle\_long \citep{kagglefn}  & 31k      & 9K      & 0.98                                                                & 0.53                                                                     & 46\%     \\
    coaid  \citep{coaid}        & 5K       & 1K      & 0.97                                                                & 0.56                                                                     & 42\%     \\
    cov19\_text \citep{cov19fn}  & 2.5K     & 0.5K    & 0.98                                                                & 0.58                                                                     & 41\%     \\
    cov19\_title \citep{cov19fn} & 2.5K     & 0.5K    & 0.95                                                                & 0.62                                                                     & 35\%     \\
    rumor \citep{covrumor}        & 4.5K     & 1K      & 0.83                                                                & 0.54                                                                     & 35\%     \\
    cq  \citep{covidcq}          & 12.5K    & 2K      & 0.71                                                                & 0.51                                                                     & 28\%     \\
    miscov19  \citep{miscov}    & 4K       & 0.6K    & 0.68                                                                & 0.50                                                                     & 26\%     \\
    covid\_fake \citep{covidfn}  & 4K       & 2K      & 0.96                                                                & 0.52                                                                     & 45\%     \\ \bottomrule
    \end{tabular}
    \end{table}

\PP{MiDAS Datasets}
We use 9 fake news datasets, shown in \cref{tab:datasets}. 
Where available, we used provided train-test splits; 
otherwise, we performed class-balanced 70:30 splits.
We performed a preliminary motivating
evaluation, shown in \cref{tab:datasets}.
Here, we have compared an oracle case to the concept
drift case.
In the oracle case, we train 9 models, one on each dataset,
and then evaluate this model on its corresponding dataset's
test set.
This is the case where the prediction data matches the 
training distribution.
In the concept drift case, we trained a model on 8 datasets,
then evaluated on the held-out dataset.
In this case, the prediction dataset, even though on the same
topic of Covid-19 fake news detection, does not match
the training distribution.
We see significant accuracy drops, between 20\% to 50\%.
This matches the observations in generalizability in \citep{kmp} and 
\citep{generalization}.

\PP{MiDAS Evaluation}
We evaluate \sys with held-one-out testing at the dataset level
similar to the generalization studies approach in \citep{kmp}. 
Our results are presented in \cref{sec:meval} 
To test \sys, we first train the encoder to learn a 
domain invariant representation with all but one dataset. 
Then we evaluate \sys' performance on classifying the 
unseen, \ie drifted, dataset.
We repeat this for each dataset in \cref{tab:datasets}.
\sys' performance is compared to Snorkel \citep{snorkel}, EEWS \citep{eews},
 an ensemble, and an `oracle' fine-tuned AlBert model 
 trained on the held-out dataset.

\PP{Ablation Study}
To further evaluate \sys' efficacy, we also conducted an 
ablation study in \cref{sec:ablation} by varying the number of training datasets, 
types of loss functions, masked language model trainig, and loss weights.

\PP{Adjustment of $m$}
For each experiment in \cref{sec:meval}, we sampled points in an $\epsilon$-ball 
around $x$, where $\epsilon$ calculated using steps \cref{sec:midasimpl}. 
We explore effects of adjusting the radius of this sampling ball by changing $m$, the the number of nearest neighbors, in \cref{sec:adjustment}. 
Specifically, we show that as the sampling/perturbation radius increases 
beyond $\epsilon$, \sys' accuracy decreases. 
Conversely, as sampling radius is reduced, \sys increases 
accuracy while sacrificing coverage.

\begin{figure}[t]
    \centering
    \begin{subfigure}{0.2\textwidth}
        \includegraphics[width=\textwidth]{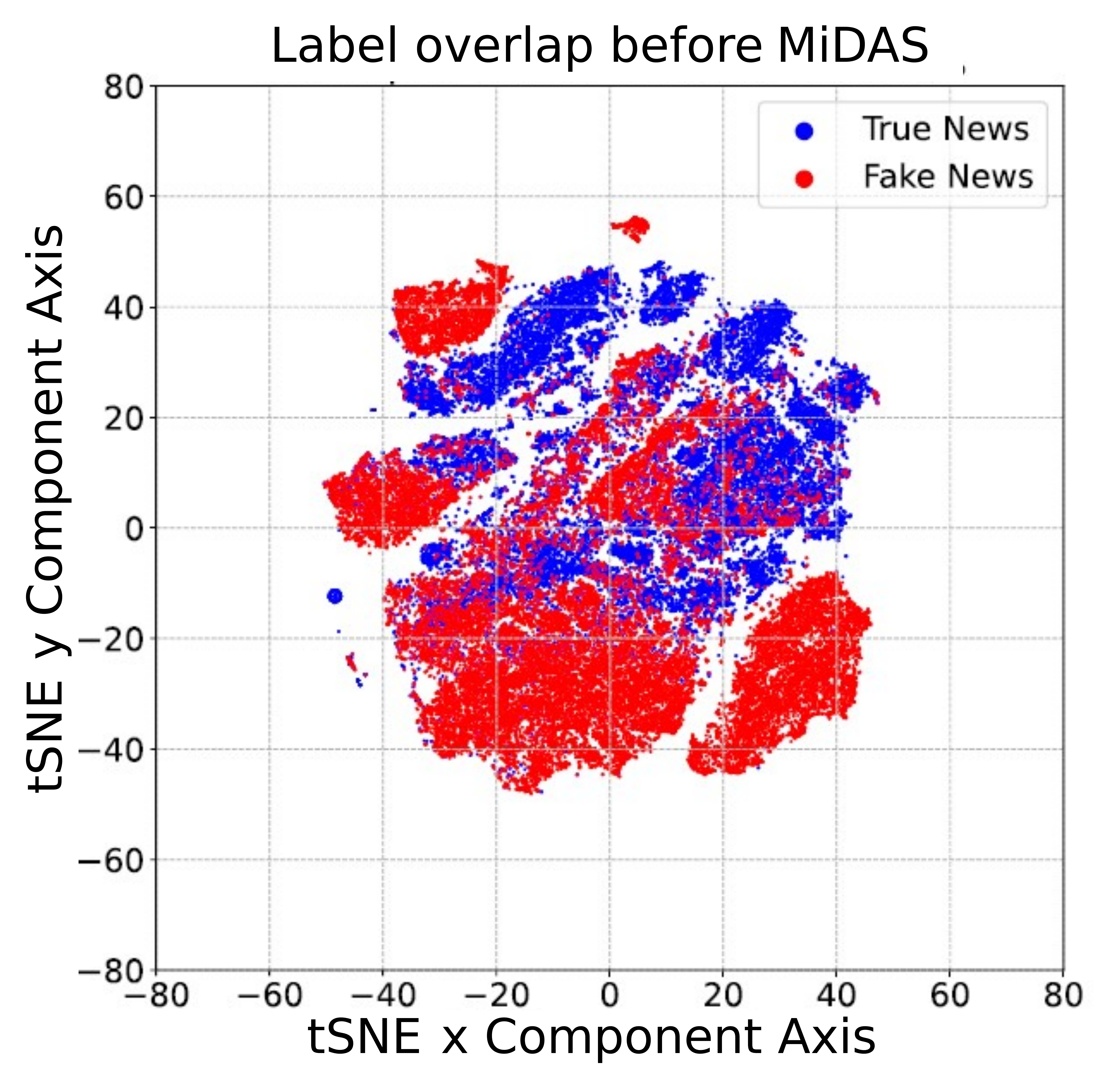}
        \caption{Overlapping clusters before MiDAS}
        \label{fig:beforemidas}
    \end{subfigure}
    \hfill
    \begin{subfigure}{0.2\textwidth}
        \includegraphics[width=\textwidth]{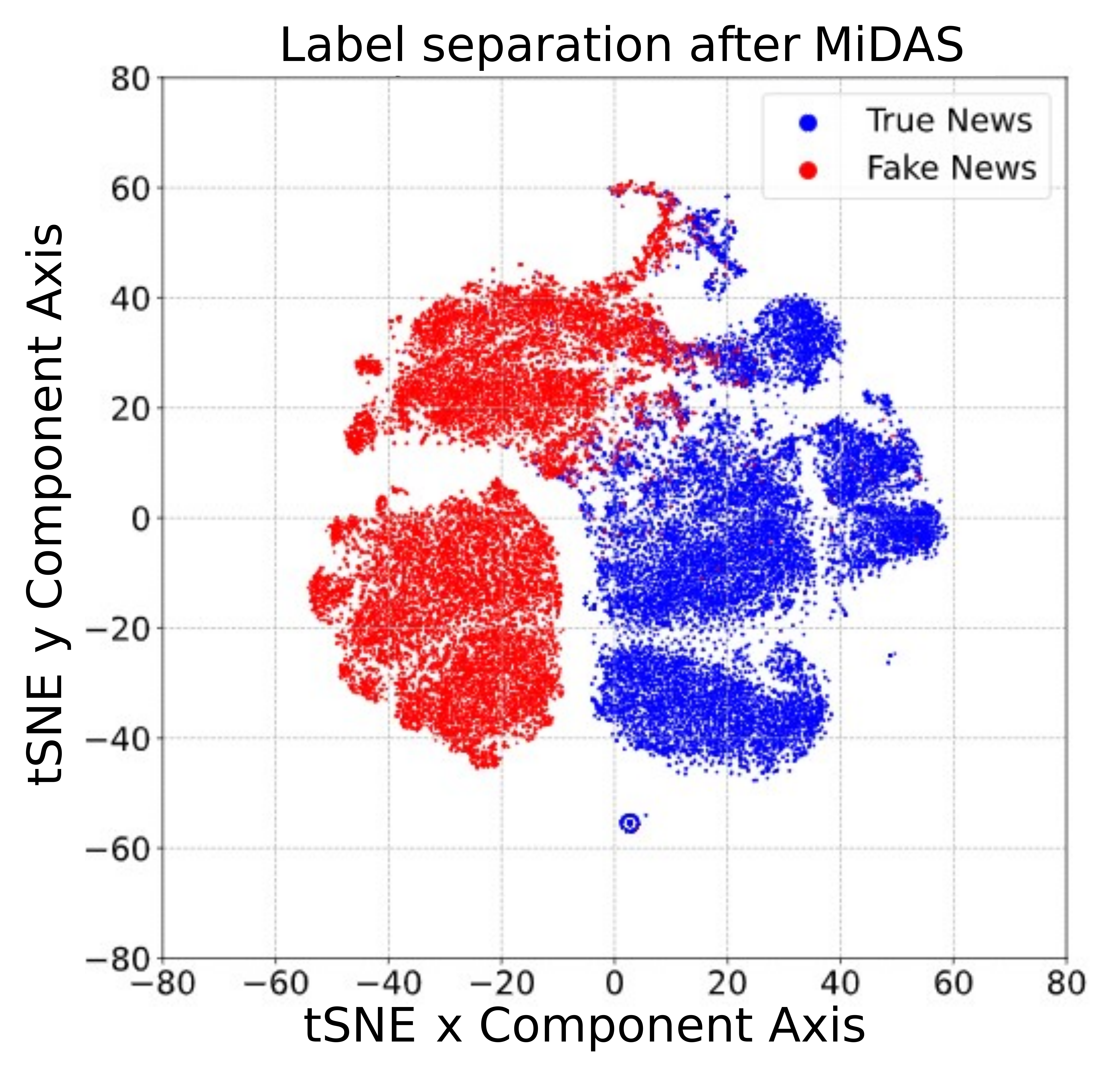}
        \caption{Domain invariance after MiDAS}
        \label{fig:aftermidas}
    \end{subfigure}
    \hfill
    \begin{subfigure}{0.28\textwidth}
        \includegraphics[width=\textwidth]{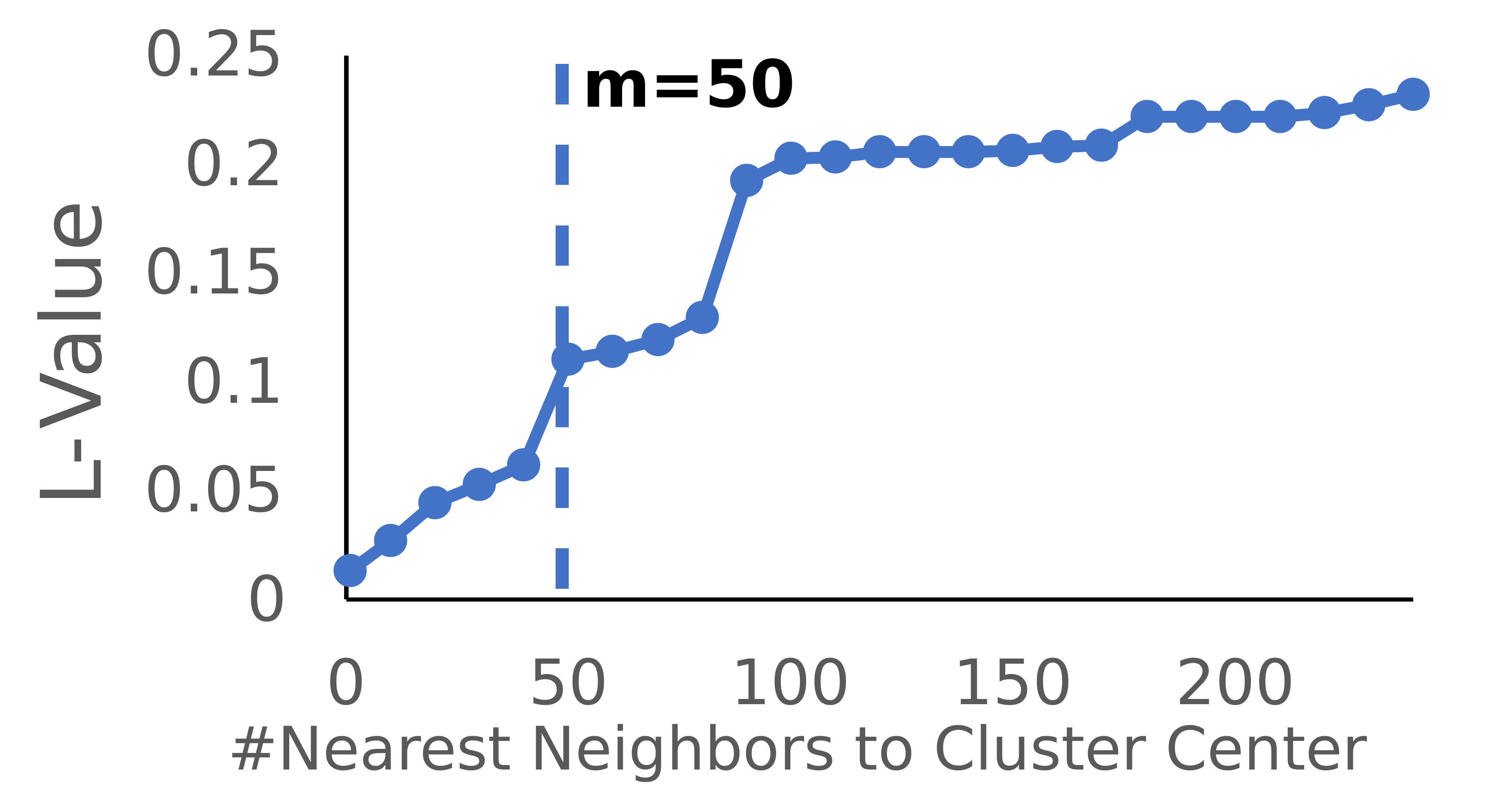}
        \caption{L-Values for different $m$ }
        \label{fig:lvals}
    \end{subfigure}
    \hfill
    \begin{subfigure}{0.28\textwidth}
        \includegraphics[width=\textwidth]{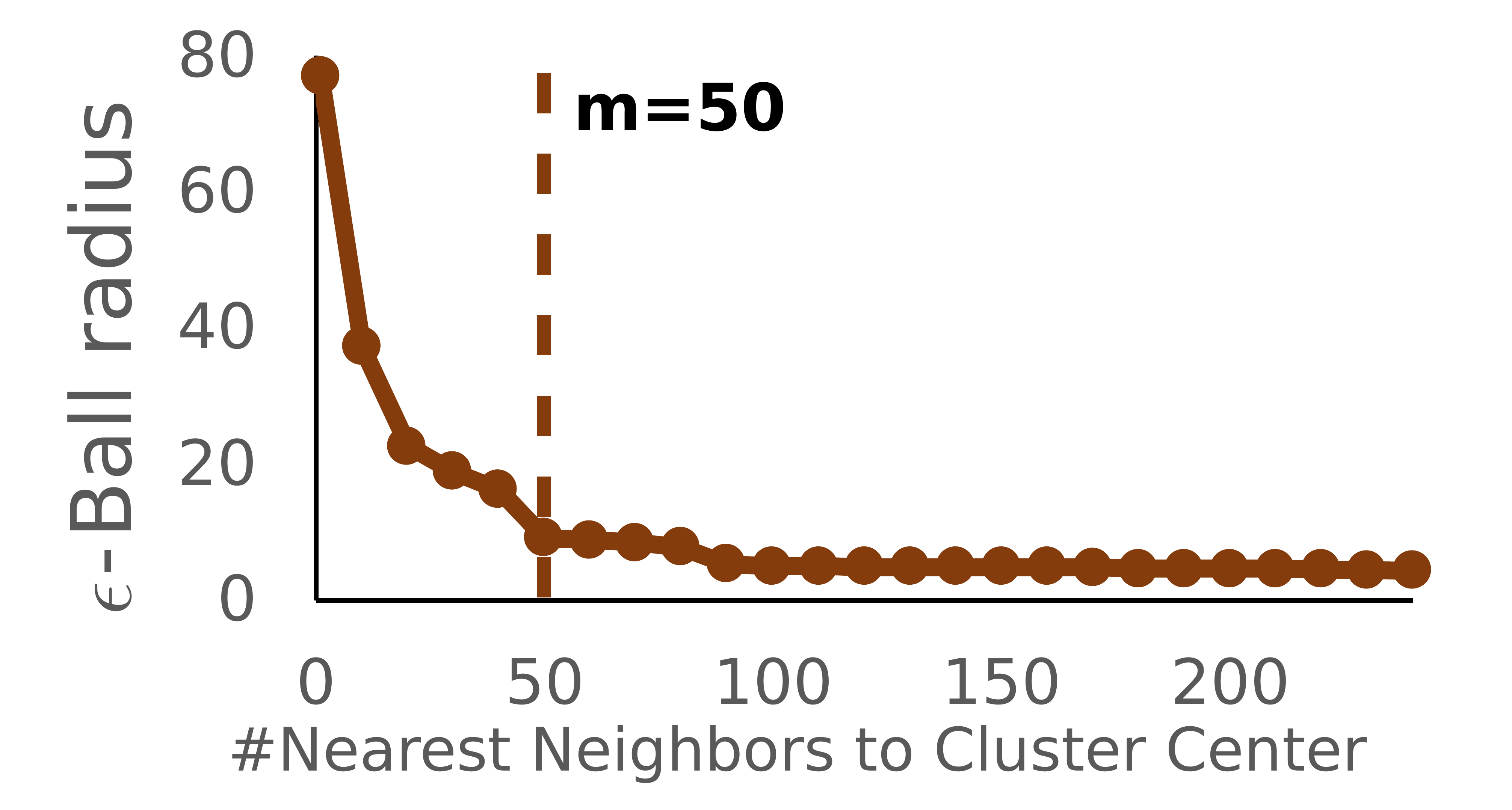}
        \caption{$\epsilon$-Ball radius for different $m$}
        \label{fig:eball}
    \end{subfigure}
            
    \caption{\textbf{MiDAS characteristics.} \cref{fig:beforemidas} shows tSNE projection of 
    all 9 datasets' pretrained BERT encoder's embeddings. \cref{fig:aftermidas} shows dataset 
    embeddings when generated with MiDAS' encoder. The label overlaps between true news and fake news 
    have been separated; in this case, enforcing domain invariance forces true and fake labels of all datasets
    to cluster. In \cref{fig:lvals} and \cref{fig:eball}, we show impact of L-Value and resulting
    $\epsilon$ calculation for different values of $m$. We select $m=50$ for remaining experiments.}
    \label{fig:graphs}
    \end{figure}

\subsection{MiDAS Evaluation}
\label{sec:meval}
We now present evaluation results for  \sys.
In each experiment, we designate a single dataset as the target dataset 
without labels, and the remaining datasets act as source domains. 
In these cases, the held-one-out dataset acts as the drifted dataset,
similar to the generalization experiments in \citep{kmp}.
We follow the steps in \cref{sec:midasimpl} to test \sys, and compare
classification accuracy to 5 approaches:
(i) an ensemble of the training models with equal weights, 
(ii) a Snorkel labeler \citep{snorkel} that treats each model as a labeling function, 
(iii) an EEWS labeler \citep{eews} that treats each model as a labeling function,
(iv) a KMP-model \citep{kmp} that uses KMeans clustering with proxies to compute overlap, 
and (v) an `oracle' AlBERT model fine-tuned on the held-out dataset.

\cref{fig:graphs} shows several characteristics of the MiDAS encoder.
\cref{fig:aftermidas} shows domain invariance in the labels.
Each point is a sample from the 9 datasets; before applying
MiDAS, there is significant label overlap because each dataset exists in separate 
domains. 
After applying MiDAS, datasets are projected to a domain invariance embedding.
This forces sampes with the same label, irrespective of source domain, to cluster 
together and reduce the label overlap observed in \citep{kmp}.

After training \sys' encoder, we need to compute the $\epsilon$ radius using 
$m$ nearest neighbors to the label cluster centers. 
We examine the impact of different $m$ values for computing $\epsilon$
in \cref{fig:lvals} and \cref{fig:eball}.
As we increase the number of neighbors used in estimating local $L$, 
the estimate for $L$ increases, indicating reduced smoothness the further we
deviate from the cluster center.
In turn, this reduces the acceptable $\epsilon$-ball radius to bound
probability of label change, per \citep{nice}, since $\epsilon = 1/L$.
A large $m$ would significantly reduce the size of the sampling $\epsilon$-ball,
and perturbations would be negligible.
A small $m$ would yield a poor estimate for local $L$ and a large sampling 
ball.
We further explore impact of changing $m$ directly on 
accuracy in \cref{sec:adjustment}. 
Here, we fix $m=50$ for remaining experiments, since we observed the
$\epsilon$-Ball radius generally stabilized around this value.
%
%
%Here, we fix $m=50$ for each experiment as a good empirical 
%trade-off between coverage and accuracy. 
%

% We show the domain invariant encoder space in \cref{fig:tsne}.
% %
% Here, we have compute tSNE over the training data embeddings before
% and after training the encoder. 
% %
% \sys projects the individual source domains to an invariant space 
% for proper $L_k$ computation.
% %
% Further, similar clusters of positive and negative samples are
% closer together.
%
\begin{table}[h]
    \caption{MiDAS Evaluation: MiDAS outperforms on generalizing to unseen, drifted data points. For each held-out dataset, given 8 fine-tuned models trained on the remaining 8 datasets, MiDAS is able to select the best-fit model for each sample. Using this best-fit model, MiDAS outperforms an equal-weighted ensemble by over 30\%.}
    \label{tab:midas}
    \small
    \begin{tabular}{@{}lcrrrrr@{}}
    \toprule
    \multicolumn{1}{c}{\multirow{2}{*}{\textbf{Dataset}}} & \textbf{Oracle Labels}         & \multicolumn{5}{c}{\textbf{No access to Labels}}                                \\ \cmidrule(l){2-7} 
    \multicolumn{1}{c}{}                                  & \multicolumn{1}{c}{FT-AlBert} & Ensemble & Snorkel & EEWS          & KMP  & \multicolumn{1}{r}{\textbf{MiDAS}} \\ \midrule
    \multicolumn{1}{l}{kaggle\_short}                   & \multicolumn{1}{c}{0.97}      & 0.57     & 0.61    & 0.67          & 0.69 & \multicolumn{1}{r}{\textbf{0.86}}  \\
    \multicolumn{1}{l}{kaggle\_long}                    & \multicolumn{1}{c}{0.98}      & 0.53     & 0.63    & 0.68          & 0.70 & \multicolumn{1}{r}{\textbf{0.71}}  \\
    \multicolumn{1}{l}{coaid}                           & \multicolumn{1}{c}{0.97}      & 0.56     & 0.64    & 0.74          & 0.81 & \multicolumn{1}{r}{\textbf{0.84}}  \\
    \multicolumn{1}{l}{cov19\_text}                     & \multicolumn{1}{c}{0.98}      & 0.58     & 0.59    & 0.68          & 0.75 & \multicolumn{1}{r}{\textbf{0.79}}  \\
    \multicolumn{1}{l}{cov19\_title}                    & \multicolumn{1}{c}{0.95}      & 0.62     & 0.69    & 0.75          & 0.61 & \multicolumn{1}{r}{\textbf{0.81}}  \\
    \multicolumn{1}{l}{rumor}                           & \multicolumn{1}{c}{0.83}      & 0.54     & 0.59    & 0.62          & 0.45 & \multicolumn{1}{r}{\textbf{0.67}}  \\
    \multicolumn{1}{l}{cq}                              & \multicolumn{1}{c}{0.71}      & 0.51     & 0.54    & 0.52          & 0.56 & \multicolumn{1}{r}{\textbf{0.57}}  \\
    \multicolumn{1}{l}{miscov19}                        & \multicolumn{1}{c}{0.68}      & 0.50     & 0.52    & \textbf{0.56} & 0.45 & \multicolumn{1}{r}{{\ul 0.54}}     \\
    \multicolumn{1}{l}{covid\_fake}                                           & 0.96                           & 0.52     & 0.56    & 0.61          & 0.50 & \multicolumn{1}{r}{\textbf{0.75}}                      \\ \bottomrule
    \end{tabular}
\end{table}

In \cref{tab:midas}, we show results of \sys compared to the 5 approaches discussed above.
In all but one of our experiments, \sys outperforms other 
labeling schemes in classifying the unseen, drifted samples. 
On average, \sys sees a 30\% increase in accuracy compared to an ensemble. 
Further, by using the training data itself to adaptively guide SM 
selection, MiDAS improves by ~21\% on Snorkel, ~10\% on EEWS and KMP.

\subsection{Ablation Study}
\label{sec:ablation}

\begin{figure} [t]
        \thisfloatsetup{floatrowsep=none}
        \begin{floatrow}
        
        \capbtabbox{%
          
        %\begin{table}[t]
        %\caption{Datasets used for our experiments. If splits were not available, we used a random class-balanced split. For Tweet datasets, sample counts are after rehydration, which removed some samples due to missing tweets.}
        %\label{tab:datasets}
        \scriptsize
        \begin{tabular}{@{}lrrrrr@{}}
            \toprule
            \multicolumn{1}{c}{\textbf{Dataset}} & MiDAS-Half & +Sources & +Center Loss & + Masking & +Weighted Loss \\ \midrule
            kaggle\_short                        & 0.56       & 0.68     & 0.75         & 0.79      & 0.86           \\
            kaggle\_long                         & 0.55       & 0.62     & 0.65         & 0.68      & 0.71           \\
            coaid                                & 0.53       & 0.71     & 0.76         & 0.79      & 0.84           \\
            cov19\_text                          & 0.54       & 0.68     & 0.72         & 0.75      & 0.79           \\
            cov19\_title                         & 0.58       & 0.71     & 0.76         & 0.78      & 0.81           \\
            rumor                                & 0.57       & 0.61     & 0.63         & 0.65      & 0.67           \\
            cq                                   & 0.52       & 0.54     & 0.55         & 0.55      & 0.57           \\
            miscov19                             & 0.51       & 0.53     & 0.53         & 0.54      & 0.54           \\
            covid\_fake                          & 0.51       & 0.59     & 0.64         & 0.69      & 0.75           \\ \bottomrule
            \end{tabular}
        %\end{table}
        }{%
        \caption{\textbf{MiDAS Ablation Study.} We examined impact of different design choices here. 
    Of note is that using masked language modeling is significantly useful in improving end-to-end accuracy. 
    Further, adding a center loss term ensures the domain-invariance representations have separable clusters, 
    as we see in \cref{fig:aftermidas}. Finally, with weighted loss, we give the discriminator 2x 
    the importance of the encoder masking loss to focus on domain invariance.}
    \label{tab:ablation}
        }
        
         \ffigbox[4cm]{%
                \centering
             \includegraphics[width=0.35\textwidth]{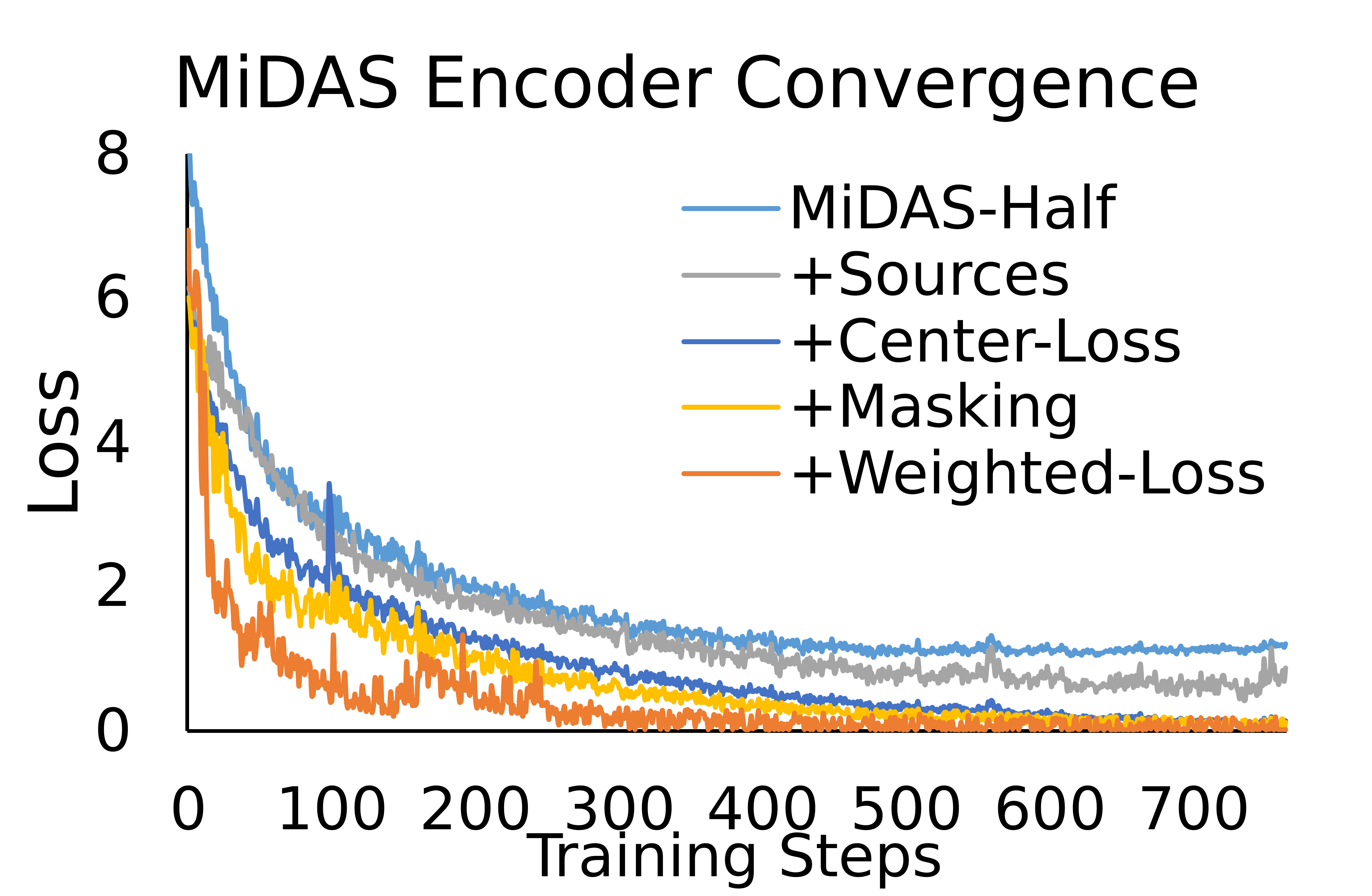}
             %\fbox{\rule[-.5cm]{0cm}{4cm} \rule[-.5cm]{4cm}{0cm}}

         }{%
         \caption{Convergence of MiDAS with each component in ablation study.}
         \label{fig:convergence}
         }
        \end{floatrow}
        \end{figure}

We evaluated the impact of several design and training choices for \sys 
in an ablation study in \cref{tab:ablation}
We use a version of MiDAS trained with half of the sources with the 
most data points for each experiment (MiDAS-Half). 
This yields near-random accuracy, since this is a modified ensemble 
on different source datasets. 
Adding the remaining sources improves MiDAS' coverage and improves accuracy by ~15\%. 
We add a center loss term \citep{centerloss} to the encoder output to encourage 
clustering on the labels between multiple sources; increasing accuracy by ~5\%. 
Next, we added language masking to the input during the encoder-decoder 
training to further fine-tune the encoder for the fake-news tasks, 
yielding a ~4\% improvement. 
Finally, we increased the weights for the discriminator loss
compared to encoder loss 
to emphasize domain invariance, yielding a 5\% improvement for MiDAS' 
accuracy on fake news detection for unseen, drifted data. 
We compare convergence for different experiments in \cref{fig:convergence}: 
the encoder converges faster in each case. 
Further, adding the center and weighted losses contribute to 
discriminator fooling and stabilizing the discriminator loss.

\subsection{Adjustment of $\epsilon$}
\label{sec:adjustment}

\begin{table}[]
    \caption{Impact of $m$ values: As we increase the number of nearest neighbors, we get a higher estimate for L, per \cref{fig:lvals}, which reduces the $\epsilon$-Ball sampling radius and relaxes threshold for model predictions. This leads to lower accuracy with higher coverage. Decreasing $m$, in turn, increases accuracy at the cost of lower coverage.}
    \label{tab:mimpact}
    \small
    \begin{tabular}{@{}lrrrrrrrr@{}}
    \toprule
    \multicolumn{1}{l}{\multirow{2}{*}{m-value}} & \multicolumn{2}{c}{kaggle\_short} & \multicolumn{2}{c}{coaid} & \multicolumn{2}{c}{rumor} & \multicolumn{2}{c}{cq} \\
    \multicolumn{1}{c}{}                         & F1-Score            & Coverage          & F1-Score       & Coverage      & F1-Score       & Coverage      & F1       & Coverage    \\ \midrule
    m=1                                          & 0.97          & 0.03              & 0.97      & 0.01          & 0.86      & 0.05          & 0.81     & 0.03        \\
    m=10                                         & 0.91          & 0.35              & 0.92      & 0.29          & 0.82      & 0.48          & 0.74     & 0.34        \\
    m=20                                         & 0.87          & 0.65              & 0.85      & 0.73          & 0.73      & 0.63          & 0.61     & 0.59        \\
    m=50                                         & 0.86          & 0.86              & 0.84      & 0.89          & 0.67      & 0.82          & 0.57     & 0.78        \\
    m=75                                         & 0.73          & 0.91              & 0.78      & 0.94          & 0.62      & 0.89          & 0.53     & 0.85        \\
    m=100                                        & 0.69          & 0.95              & 0.73      & 0.98          & 0.59      & 0.97          & 0.51     & 0.91        \\
    m=150                                        & 0.57          & 0.99                & 0.56      & 1.00           & 0.54      & 1.00            & 0.43     & 0.95           \\ \bottomrule
    \end{tabular}
    \end{table}

Finally, we investigate $\epsilon$ with respect to $m$, which was fixed at $m$=50.
For these experiments, we investigated increasing and 
decreasing $m$ to, respectively, increase and decrease $\epsilon$.
Increasing the neighbors increases the computed $L$, since we 
are using points further from the smooth cluster center. 
In turn, this reduces the sampling $\epsilon$-ball, so 
the perturbations we apply will be smaller, and 
in some cases, negligible.
Furthermore, threshold value of $L$ needed to 
accept an SM's prediction is higher (since it is
$1/\epsilon$), so MiDAS tolerates lower smoothness for
each model, and accepts predictions from more models,
resulting in higher 
coverage and lower overall accuracy.
On the other hand, using fewer neighbors means larger
sampling ball and smaller threshold for acceptance.
It is more likely for perturbed samples to be further away,
yielding a higher value of L unless a corresponding model
is especially smooth around that point.
This would, as a result, reduce coverage, but increase accuracy.

We show this in \cref{tab:mimpact} for several values of $m$
across 4 of our datasets.
Using only the nearest neighbor yields minimal coverage.
As we increase the $m$, coverage increases significantly, 
and accuracy approaches ensemble accuracy. 
Conversely, as we reduce $m$, we also reduce $L$ and 
consequently, the smoothness threshold to accept a prediction. 
This increases accuracy, since \sys rejects predictions 
that do not satisfy the threshold. 
However, coverage decreases as well: we show in 
\cref{tab:mimpact} fewer unseen samples from the target domain can be 
labeled as we decrease $m$.
We also see that $m$ can have outsized impact on accuracy as well:
`coaid' f1 scores drop from 0.73 to 0.56, even though coverage
increases only slightly, from 0.98 to 1.0 when we increase $m$ from
100 to 150.
This occurs because at $m=150$, MiDAS' relaxed thresholds allow
poorer models to provide predictions as well, reducing accuracy
in the final averaged result.

%\subsection{Advantages of MiDAS}

\subsection{Limitations and Future Directions}
\label{sec:limitations}
We tested MiDAS in the scenario where fine-tuned models already
exist.
This constrains the MiDAS encoder, which must also
train a decoder so match the inputs of the fine-tuned models
that expect tokenized input.
A more flexible approach would deploy models and MiDAS
together, with each fine-tuned model
directly trained with the MiDAS encoder.
This would improve both training time, convergence, 
as well as accuracy, since each model would 
directly use the MiDAS generated encodings,
instead of domain-specific reconstructions.

Furthermore, we selected $m$ using empirical observations. 
However, there can be technically grounded approaches as well, 
such as using the high-density bands from \citep{odin, trust}. 
In these cases, the $\alpha$-high density region of each 
cluster can be used to estimate a good $m$. 
We leave further exploration of $m$ as well as
integration of fine-tuned models into the encoder
training framework to future work.
\section{Conclusion}
\label{sec:conc}
We have presented \sys, a system for adaptively selecting 
best-fit model for a set of samples from drifting distributions.
\sys uses a domain-invariance embedding to estimate local
smoothness for fine-tuned models around drifting samples.
By using local smoothness as a proxy for accuracy and 
training data relevancy, \sys improves on 
generalization accuracy across 9 fake news datasets.
With \sys, we can detect COVID-19 related fake news with 
over 10\% accuracy improvement over weak labeling approaches.
We hope \sys will lead further exploration into the tradeoff between
generalizability and fine-tuning, as well as research into
mitigating generalization difficulties of pre-trained models.

\newpage

%\section*{References}
\bibliographystyle{plainnat}
\bibliography{main}

\newpage

%%%%%%%%%%%%%%%%%%%%%%%%%%%%%%%%%%%%%%%%%%%%%%%%%%%%%%%%%%%%

%\input{checklist}

%\input{supplementary}

\end{document}